\def\year{2023}\relax
\documentclass[letterpaper]{article} % DO NOT CHANGE THIS
\usepackage[submission]{aaai23}  % DO NOT CHANGE THIS
\usepackage{times}  % DO NOT CHANGE THIS
\usepackage{helvet}  % DO NOT CHANGE THIS
\usepackage{courier}  % DO NOT CHANGE THIS
\usepackage[hyphens]{url}  % DO NOT CHANGE THIS
\usepackage{graphicx} % DO NOT CHANGE THIS
\usepackage{natbib}  % DO NOT CHANGE THIS AND DO NOT ADD ANY OPTIONS TO IT
\usepackage{caption} % DO NOT CHANGE THIS AND DO NOT ADD ANY OPTIONS TO IT
\usepackage{multirow}
\usepackage{amsmath}
\usepackage{algorithm}
\usepackage{xspace}
\usepackage{algorithmic}
\usepackage{amsmath,amssymb,amsfonts}
\usepackage{algorithmic}
\usepackage{graphicx}
\usepackage{hhline}
\usepackage{textcomp}
\usepackage{xcolor}
\usepackage{comment}
\usepackage{times}
\usepackage{romannum}
\usepackage{latexsym}
\usepackage[T1]{fontenc}
\usepackage[utf8]{inputenc}
\usepackage{graphicx}
\usepackage{amsmath}
 
\usepackage{amssymb}

\usepackage{multirow}
\usepackage{tabto}
\usepackage{booktabs} % For formal tables
\usepackage{algorithmic}
\usepackage{algorithm}
\usepackage{graphicx}
\usepackage{subcaption}
\usepackage{todonotes}
\usepackage{enumitem}

\usepackage{newfloat}
\usepackage{listings}
\DeclareCaptionStyle{ruled}{labelfont=normalfont,labelsep=colon,strut=off} % DO NOT CHANGE THIS
\floatstyle{ruled}
\newfloat{listing}{tb}{lst}{}
\floatname{listing}{Listing}
\title{HyperHawkes: Hypernetwork based Neural Temporal Point Process}

\author{Manisha Dubey}
\author{
    Manisha Dubey,
    P.K. Srijith,
    Maunendra Sankar Desarkar
    \\
    cs17resch11003@iith.ac.in, srijith@cse.iith.ac.in, maunendra@cse.iith.ac.in \\
  Indian Institute of Technology Hyderabad, India  
}
\affiliations{
    \textsuperscript{}Computer Science and Engineering\\
    Indian Institute of Technology Hyderabad, India\\
    cs16resch11004@iith.ac.in,srijith@cse.iith.ac.in\\
    }

\begin{document}
\maketitle

\begin{abstract}
 Temporal point process serves as an essential tool for modeling time-to-event data in continuous time space. Despite having massive amounts of event sequence data from various domains like social media, healthcare etc., real world application of temporal point process faces two major challenges: 1) it is not generalizable to predict events from unseen sequences in dynamic environment 2) they are not capable of thriving in continually evolving environment with minimal supervision while retaining previously learnt knowledge. To tackle these issues, we propose \textit{HyperHawkes}, a hypernetwork based temporal point process framework which is capable of modeling time of occurrence of events for unseen sequences. Thereby, we solve the problem of zero-shot learning for time-to-event modeling. We also develop a hypernetwork based continually learning temporal point process for continuous modeling of time-to-event sequences with minimal forgetting. In this way, \textit{HyperHawkes} augments the temporal point process with zero-shot modeling and continual learning capabilities. We demonstrate the application of the proposed framework through our experiments on two real-world datasets. Our results show the efficacy of the proposed approach in terms of predicting future events under zero-shot regime for unseen event sequences. We also show that the proposed model is able to predict sequences continually while retaining information from previous event sequences, hence mitigating catastrophic forgetting for time-to-event data. 
\end{abstract}

\section{Introduction}

Various applications in daily life like earthquake occurrences, social networks, financial transactions, user activity logs etc. are associated with collection of discrete asynchronous events where event occurrences are represented with timestamps.  Each event sequence, consisting of a series of timestamps, is associated with a separate entity. For example, in social media, each user can be associated with the time of posting a tweet, and each tweet can be viewed as an event. Similarly, in financial transactions, each stock can be associated with the time of buy-sell order. The ability to model such sequences is of vital importance to create intelligent systems. These sequences often contain rich information, which can predict the future evolution of the sequences.

A principled mathematical framework to model such sequences in continuous time space is temporal point process \cite{valkeila2008introduction}. Hawkes process~\cite{hawkes1971spectra}, a self exciting point process, has a rich literature in terms of theoretical importance and has been widely used in a wide array of practical applications like epidemic modeling\cite{paper18}, earthquake prediction\cite{paper20}, financial modeling\cite{bacry2015hawkes}, crime prediction\cite{paper19} etc.    Recent works improve the performance of standard classic Hawkes process by considering neural networks for modeling such event sequences~\cite{mei2016neural, du2016recurrent, xiao2017modeling, omi2019fully}. Neural Hawkes processes have proved to learn complex dependencies as against their classical counterparts. The Neural Hawkes process is one of the cornerstones of recent progress in time-to-event modeling.

Despite having improved performance, the neural Hawkes process is challenged by two potential limitations on the practical side. 
%First, the deployment of the neural Hawkes process suffers from a cold-start problem. 
Neural Hawkes process typically needs to be trained on a large time-to-event dataset for the specific domain or entity. This restricts the prediction for a new and unseen entity with limited or no data and can be very crucial for certain applications. Moreover, the process of data acquisition for the time of occurrence of events for a new sequence is expensive. Besides, such a process can be time consuming for some sequences which may have low frequency of occurrence, and hence may take a long time to produce massive amounts of data which will be used to predict future time of occurrences. Secondly, real-world event occurrences happen sequentially in continuous streams. Therefore, a realistic and challenging problem is to continually learn time-to-event models in an ever-changing environment while retaining previous learnt knowledge.  

Motivated by the above limitations, we consider a practical and under-explored setting for time-to-event modeling, called zero-shot event modeling. We also consider a continual learning setup  where time-to-event prediction tasks arrive sequentially in an online manner. We aim to develop neural Hawkes process models which could generalize to time-to-event prediction tasks with no data and can continually learn while retaining previous knowledge. To this end, we introduce \textit{HyperHawkes}, a hypernetwork based Hawkes process to generate sequence-specific parameters for the neural Hawkes process. Hypernetwork is essentially a metanetwork which can generate parameters for the  neural Hawkes process network for modeling continuous events.  By employing hypernetwork based learning, we improve the model's generalization ability to predict unseen sequences using the sequence descriptors. By incorporating descriptor-conditioned hypernetwork, we enable learning at the level of time-to-event sequence by learning event-sequence-specific parameters, hence being able to predict unseen sequences with the help of a descriptor. For a more pragmatic setup, we augment our model to consider continually arriving sequences where each sequence can be considered as a separate task. For continually learning the event sequences, we recast the descriptor-conditioned hypernetwork to include a hypernetwork output regularizer. This regularizer will penalize the changes in previously learnt parameters, hence retaining previously learnt time-to-event modeling capabilities.  We provide two variations to the proposed approach, allowing to encompass 1) zero-shot modeling 2) continual learning capabilities within the framework of neural Hawkes process. To the best of our knowledge, there is no prior work on zero-shot or continually learning time-to-event modeling.

%  In this work, we address the problem of reducing the cost of modeling event sequences using the neural Hawkes process in the industrial practice of event modeling. We show that it is feasible to train a neural Hawkes process on a diversified source dataset and deploy it on a target dataset in a zero-shot regime. To the best of our knowledge, this is a novel problem addressed on event-modeling set-up. 

% We believe that addressing this practical problem provides clues to some fundamental questions on the nature of event occurrence in real world scenarios. Through our experiments, we want to enquire if we can learn something general which is common across sequences about forecasting and transfer this knowledge across different sequences. Moreover, methods designed for classification problems, such as prototypical networks cannot be directly applied to event modeling tasks. We introduce a framework where we predict events for unseen sequences by incorporating sequence descriptions and using a sequence-level training procedure. We introduce \textit{HYPERHAWKES}, a hypernetwork based Hawkes process to dynamically generate sequence-specific parameters from sequence description. Such formulations enable learning at task level in contrast with the event level. 

Our contributions can be summarized as follows:
\begin{itemize}
    \item We propose two novel problems of zero-shot learning and continually learning from the paradigm of time-to-event modeling. 
    \item We propose \textit{HyperHawkes}, a descriptor-conditioned hypernetwork based neural Hawkes process which can generate event sequence specific parameters, hence learning at the level of sequence.  We present two variants of \textit{HyperHawkes} considering architecture of neural Hawkes process.
    % \item We propose two variants of \textit{HyperHawkes} considering architecture of neural Hawkes process.
    \item The proposed methods can be used for predicting time of occurrences of unseen sequences, hence performing zero-shot time-to-event modeling.
    \item We augment the model with continual learning abilities by employing hypernetwork based regularization parameter, hence avoiding catastrophic forgetting for successively appearing time-to-event sequences.
    \item We present an experimental  setup for evaluating zero shot learning and continual learning for time-to-event modeling.  We demonstrate the effectiveness of the proposed models on these setups for two real-world datasets.   
\end{itemize}

% We apply \textit{HYPERHAWKES} to two datasets: Yelp and Meme. We demonstrate the effectiveness of the proposed framework. Since this is a novel problem, we also propose baselines for the given problem.  

%  We perform this by considering sequence description as a part of the model.

\section{Related Work}
\subsection{Hawkes Process}
Hawkes process~\cite{hawkes1971spectra} is a point process \cite{valkeila2008introduction} with self-triggering property. 
i.e occurrence of previous events trigger occurrences of future events. 
Hawkes process has been used in earthquake modeling~\cite{paper20}, crime forecasting~\cite{paper19}, social media~\cite{rizoiu2017tutorial} finance~\cite{bacry2015hawkes,embrechts2011multivariate} and epidemic forecasting~\cite{paper18,chiang2021hawkes}. 
They provide a solid mathematical framework for modeling event sequences. Earlier works on point process modeling specify a parametric form for the  intensity function characterizing the point process. However, parametric models may not be capable of capturing the complex event dynamics. 
%~\cite{du2016recurrent} has proposed the use of recurrent neural networks to learn the compact representation of history. Further, the conditional intensity function is modeled as a function of the hidden state of RNN. 
To address this, several research works were proposed~\cite{du2016recurrent,mei2017neural,omi2019fully,zuo2020transformer} where the intensity function is modeled using neural networks, which are better at learning  complex event dynamics. Recently, \cite{zuo2020transformer} and \cite{zhang2020self} proposed to use positional encodings in transformer language models \cite{vaswani2017attention} to model point processes. There are some efforts for learning from small data for Hawkes process~\cite{xie2019meta, salehi2019learning}. However, they are based on the statistical Hawkes process model (not the neural Hawkes process) and are not applicable to a zero shot learning setting.  

\subsection{Hypernetwork, Continual Learning and ZSL}

\paragraph{\textbf{Zero-Shot Learning:}} 
ZSL~\cite{palatucci2009zero,lampert2009learning} aims to predict classes which are not in training samples. Such classes are known as unseen classes and classes which are in training samples are known as seen classes. A few methods to address zero-shot learning is through mapping function \cite{frome2013devise}, generative model \cite{felix2018multi} and graph neural network \cite{wang2018zero}. Huge literature has addressed the problem of zero-shot learning across various domains of vision, natural language processing tasks such as text classification, relation extraction etc.

\paragraph{\textbf{Continual Learning:}} aims to create a learning paradigm which is able to model a stream of tasks while avoiding catastrophic forgetting. Different techniques \cite{kirkpatrick2017overcoming,li2017learning,lopez2017gradient,von2019continual}  have been proposed in this regard by consolidating knowledge either in various spaces like data, weight or meta space.
% \cite{kirkpatrick2017overcoming} introduced a regularization based technique called Elastic Weight regularization (EWC) which penalizes drastic change in the parameters having large influence on prediction. \cite{li2017learning} presents Learning without Forgetting, a knowledge distillation based method which preserves knowledge from previous tasks. Another common way to avoid catastrophic forgetting is Gradient Episodic Memory \cite{lopez2017gradient} which stores a limited number of samples to retrain. 
\cite{von2019continual} performs continual learning through hypernetwork which learns task conditioned weights of base model. 
\paragraph{
\textbf{Hypernetworks:}} They have been introduced as a metanetwork which can generate weights for another network~\cite{ha2017hypernetworks}. It is used for various tasks like meta learning \cite{zhao2020meta}, neural architecture search \cite{zoph2016neural}, natural language understanding~\cite{he2022hyperprompt} etc.

Despite having several literature in all these domains, to the best of our knowledge, there is no work in the intersection of zero-shot learning and time-to-event modeling. Also, there is no effort along the lines of continual learning for time-to-event modeling. Therefore, we address novel and essential problems in this direction which can benefit several applications. 

\section{Preliminary}

\subsection{Problem Definition}
\begin{itemize}
    \item \textit{Zero-shot Learning for time-to-event modeling:} 
    Assume we are given a collection of N seen sequences $\mathcal{D}^S = \{(\mathcal{T}^1, \mathbf{d}^1), (\mathcal{T}^2, \mathbf{d}^2),..., (\mathcal{T}^N, \mathbf{d}^N \}$ where
    $\mathbf{d}^i$ represents the  descriptor of the $i^{th}$ sequence or meta-information  
    and $\mathcal{T}^i$ represents the times  of occurrence of $n^i$ events in the $i^{th}$ sequence,
    i.e. $\mathcal{T}^i = \{t^i_j\}_{j=1}^{n^i}$. Our goal is to predict time of event occurrences for $\bar{N}$ unseen sequences $\mathcal{D}^U = \{(\mathcal{T}^1, \mathbf{d}^1), (\mathcal{T}^2, \mathbf{d}^2), ..., (\mathcal{T}^{\bar{N}}, \mathbf{d}^{\bar{N}} \}$ with the help of the sequence descriptor. 
    
    \item \textit{Continual learning for time-to-event modeling:} Assume we are given a collection of N sequences $\mathcal{D} = \{(\mathcal{T}^1, \mathbf{d}^1), (\mathcal{T}^2, \mathbf{d}^2),  ..., (\mathcal{T}^N, \mathbf{d}^N\}$ where $\mathbf{d}^i$ represents the sequence descriptor and $\mathcal{T}^i$ represents the times  of occurrence of $n^i$ events in the $i^{th}$ sequence, i.e. $\mathcal{T}^i = \{t^i_j\}_{j=1}^{n^i}$ and we assume these sequences arrive one after the other in the order of their index. Our goal is to continually learn the sequences while avoiding catastrophic forgetting from the previous sequences. So, we aim to learn a NHP model which will be able to predict the future event occurrences in all the sequences $(\mathcal{T}^i, \mathbf{d}^i)$ where $i \leq j$ after training on the $j^{th}$ sequence $(\mathcal{T}^j, \mathbf{d}^j)$.   
\end{itemize}

\subsection{Hawkes Process}
Point processes are useful to model the distribution of points over some space and are defined using an underlying intensity function.  
A Hawkes process \citep{hawkes1971spectra} is a point process with self-triggering property i.e occurrence of previous events trigger occurrences of future events. 
% Hawkes processes are powerful tools for modeling the mutual-excitation phenomena commonly observed in social networks, quantitative finance, and
% healthcare records. 
Conditional intensity function for univariate Hawkes process at time $t^i_j$ for the $i^{th}$ sequence is defined as
\setlength{\abovedisplayskip}{0pt}
\setlength{\belowdisplayskip}{0pt}
\setlength{\abovedisplayshortskip}{0pt}
\setlength{\belowdisplayshortskip}{0pt}
\begin{equation}
  \lambda(t^i_j) = \mu_i + \sum_{k=1}^{j-1} k(t^i_j - t^i_k)
  \end{equation}
 where 
 $\mu_i$ is the base intensity function and $k(\cdot)$ is the triggering kernel function capturing the influence from previous events. The summation represents  the effect of all events prior to time $t^i_j$ which will contribute to computing the intensity at time $t^i_j$. 
%  The intensity function represents the instantaneous occurrence of an event at any time. 
%  Generally, exponentially decaying function is used as a triggering Kernel which models higher influence from recently occurring events.  
The probability density function at time $t^i_j$ given the past event times as $\{t^i_1,t^i_2,\ldots,t^i_{j-1}\}$, is obtained as follows:
\setlength{\abovedisplayskip}{0pt}
\setlength{\belowdisplayskip}{0pt}
\setlength{\abovedisplayshortskip}{0pt}
\setlength{\belowdisplayshortskip}{0pt}
\begin{equation}
p(t^i_j|t^i_1,t^i_2,\ldots,t^i_{j-1}) = \lambda(t^i_j) \exp \left\{ -\int_{t^i_{j-1}}^{t^i_j} \lambda(t)dt \right\}, \label{eq:pdf_i}
\end{equation}
where the exponential term in the right-hand side represents the probability that no events occur in $[t^i_{j-1},t^i_j)$.
% The probability density function that an event sequence $\{t^i_j\}_{j=1}^{n^i}$ is observed is then obtained as follows:
% \setlength{\abovedisplayskip}{0pt}
% \setlength{\belowdisplayskip}{0pt}
% \setlength{\abovedisplayshortskip}{0pt}
% \setlength{\belowdisplayshortskip}{0pt}
% \begin{equation}
% p(\{t^i\}_{j=1}^{n^i}) = \prod_{j=1}^{n^i} \lambda(t^i_j) \exp \left\{- \int_{0}^T \lambda(t) dt\right\}. \label{eq:pdf}
% \end{equation}
% where $[0,T]$ is the time interval of observations.  
% Hawkes process has been used in earthquake modelling \citep{paper20}, crime forecasting \citep{paper19} and epidemic forecasting \citep{paper18}.
 
\begin{figure*}
\begin{subfigure}{0.45\textwidth}
\centering
\includegraphics[height=0.52\textwidth]{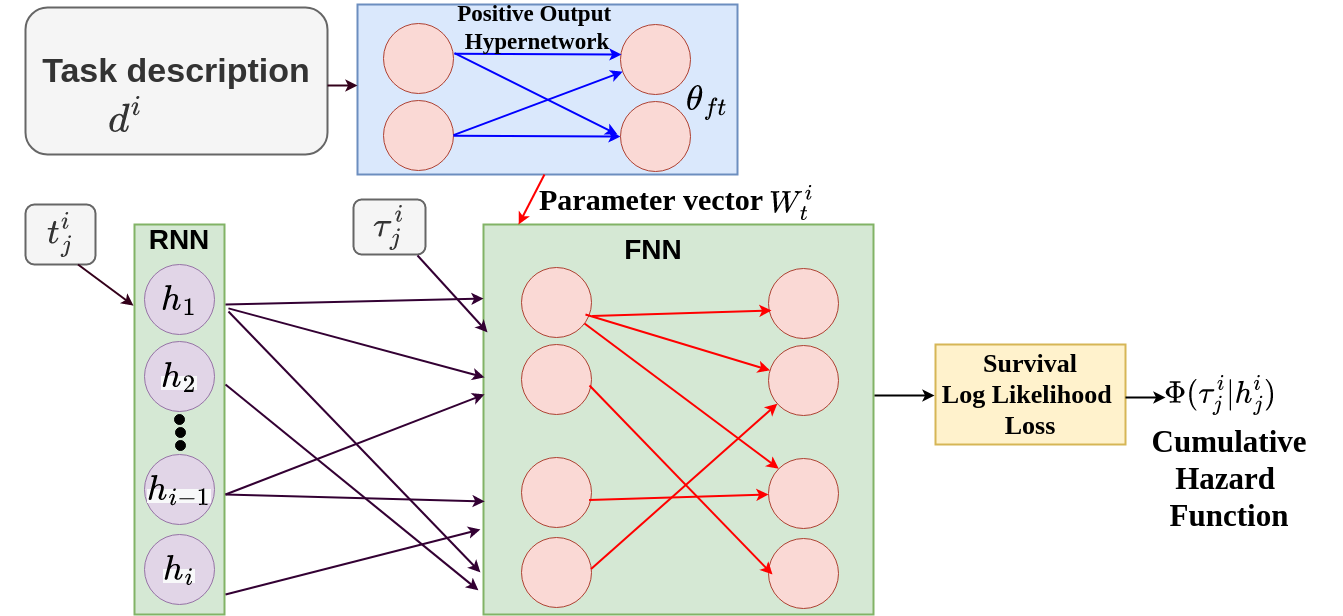}
\caption{}
\label{fig:framework_hyper_rnn}
\end{subfigure}
\quad \quad \quad
\begin{subfigure}{0.48\textwidth}
\centering
\includegraphics[ height=0.54\textwidth]{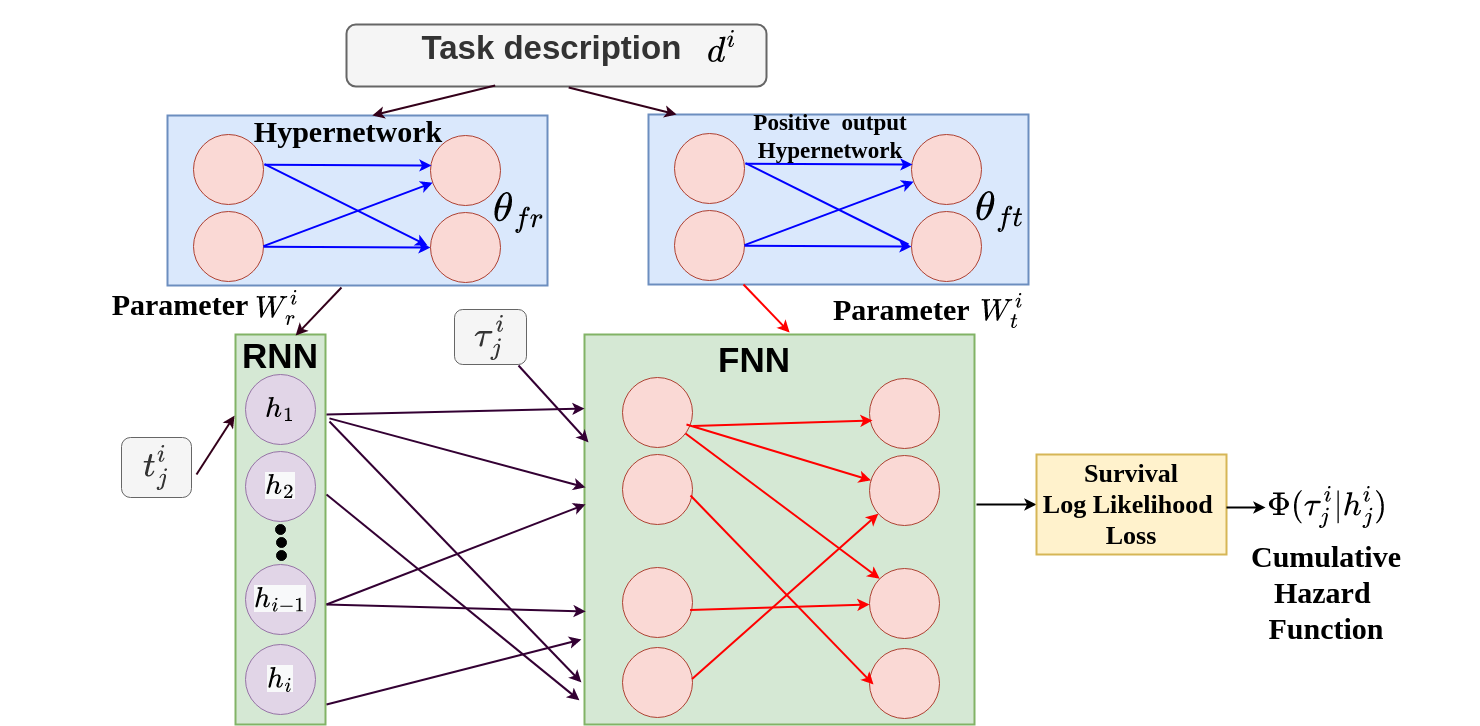}
\caption{}
\label{fig:framework_hyper_rnn_chfn}
\end{subfigure}
	\caption{Framework for the proposed model HyperHawkes \ref{fig:framework_hyper_rnn})  shows hypernetwork being used only for FNN and is called HyperHawkes-FNN, \ref{fig:framework_hyper_rnn_chfn})  shows hypernetworks used for FNN and RNN and is  called HyperHawkes-FNN-RNN}
    \label{fig:overview}
\end{figure*}

\subsection{Neural Hawkes process}
% A major characteristic of the Hawkes process~\cite{hawkes1971spectra} is   the conditional intensity function which conditions the next event occurrence based on the history of events. 
Standard Hawkes process assumes a parametric form for the intensity function which is not generalizable to every event prediction problem. The influences between the events can be complex and need not be exponentially decaying. Various recent works introduced neural Hawkes processes~\cite{du2016recurrent,mei2016neural,omi2019fully} which models the intensity function as a nonlinear function of history  using a neural network. The central idea of these works is to use recurrent neural networks (RNN) to model intensity function which  captures the influence of past events. So, conditional intensity function is modeled as: 
\[
\lambda(t^i_j) = f(\mathbf{h}_j^i)
\]
where $\mathbf{h}_j^i$ represents a hidden state updated using RNN and $f(\cdot)$ is a positive valued function ensuring positivity of the intensity function.

\section{Proposed Model}
% - We propose a decomposable neural hawkes process with the help of hypernetwork. 

% - Two-stage approach where task attributes are learnt in the first stage and events are predicted in next stage. 

% - We propose a three-stage approach for event prediction. In the first stage, we learn the task embeddings. In the second stage, we generate weights for the neural hawkes process. And the last stage comprises of predicting event using the generated weights. 

% - Name hypernetwork layer as adapter (or other synonym), mention number of layers in each adapter. 

% - Hypernetwork refers to a framework in which a neural network is trained to predict weights of another network. 

We propose \textit{HyperHawkes}, a hypernetwork based neural Hawkes process for time-to-event modeling. For time-to-event modeling, we consider the neural Hawkes process (NHP) \cite{omi2019fully} as the base model. Integrating Hypernetwork with neural Hawkes process, we introduce descriptor-conditioned hypernetwork to generate weights for each sequence which can perform time-to-event modeling. The descriptor-conditioned hypernetwork learns separate weights of NHP for each sequence. We leverage this framework for zero-shot event modeling where the hypernetwork produces weights for unseen tasks using the sequence descriptor and NHP predicts future events. Inspired by \cite{von2019continual}, we also use this framework for continual learning of tasks by using a hypernetwork based regularizer. We discuss each of these pieces in detail in further subsections.

\subsection{Base Model: Neural Hawkes Process}
\label{sec:nhp}
In particular we employ neural Hawkes process \cite{omi2019fully} as base model for time-to-event modeling. It uses a combination of recurrent neural network and feedforward neural network to model the intensity function.
% This  allows the intensity function to take any functional form depending on  data and help in better generalization performance.  
% Let's denote the  neural network function as $f^W(\cdot)$ with $W$ representing the model parameters (weight vectors). 
%Please note that $W$ represents parameters associated with a single neural network.
We represent history by using hidden representations generated by recurrent neural networks (RNNs) at each time step. The hidden representation $\boldsymbol{h}^i_j$ at time $t^i_j$ is obtained as
% \setlength{\belowdisplayskip}{0pt} \setlength{\belowdisplayshortskip}{0pt}
% \setlength{\abovedisplayskip}{0pt} \setlength{\abovedisplayshortskip}{0pt}
% \begin{equation}
% \boldsymbol{h}_i = f ( W^h \boldsymbol{h}_{i-1} + W^t {\tau}_i + \boldsymbol{b}^h ), \label{eq:rnn}
% \end{equation}
\setlength{\abovedisplayskip}{0pt}
\setlength{\belowdisplayskip}{0pt}
\setlength{\abovedisplayshortskip}{0pt}
\setlength{\belowdisplayshortskip}{0pt}
\begin{equation}
\boldsymbol{h}^i_j = RNN(\tau^i_j, \boldsymbol{h}^i_{j-1}; W^i_r ) = \sigma(\tau^i_j V^i_r +  \boldsymbol{h}^i_{j-1} U^i_r + b^i_r)
\label{eq:rnn}
\end{equation}
where $\tau^i_j = t - t^i_j$ and $W^i_r$ represents the parameters associated with RNN for the $i^{th}$ sequence such as  input weight matrix $V^i_r$, recurrent weight matrix $U^i_r$, and and bias $b^i_r$.  $\boldsymbol{h}^i_j$ is obtained by repeated  application of the RNN block on a sequence formed from  previous $M$ inter-arrival times.
%$\tau_{i-(M-1)}, \ldots, \tau_{i-1},\tau_{i}$.
This is used as input to a feedforward neural network to compute the intensity function (hazard function) and consequently the cumulative hazard function for computing the likelihood of  event occurrences.
%As proposed by~\cite{omi2019fully}, a feedforward neural network which takes this hidden representation as input can be used to model the intensity function (hazard function) and consequently the cumulative hazard function for computing the likelihood of occurrence of events. 
In the proposed model, input to the feed-forward neural network is \Romannum{1}) the hidden representation generated from RNN \Romannum{2}) elapsed time from the most recent event. We model the conditional intensity as a function of the elapsed time from the most recent event as- 
% $\lambda(t-t_i|\boldsymbol{h}_i, t)$,
\setlength{\abovedisplayskip}{0pt}
\setlength{\belowdisplayskip}{0pt}
\setlength{\abovedisplayshortskip}{0pt}
\setlength{\belowdisplayshortskip}{0pt}
\begin{equation}
\lambda(t|H^i_t) = \lambda(t-t^i_j|\boldsymbol{h}^i_j)
\end{equation}
where $\lambda(\cdot)$ is a non-negative function referred to as a hazard function.
 Therefore, we define cumulative hazard function in terms of inter-event interval $\tau^i_j = t - t^i_{j}$ as $\Phi(\tau^i_j|\boldsymbol{h}^i_j) = \int_{0}^{\tau^i_j}\lambda(s|\boldsymbol{h}^i_j)ds$. Cumulative hazard function is modeled using a  feed-forward  neural network (FNN)
 \setlength{\abovedisplayskip}{0pt}
\setlength{\belowdisplayskip}{0pt}
\setlength{\abovedisplayshortskip}{0pt}
\setlength{\belowdisplayshortskip}{0pt}
\begin{equation}
\Phi(\tau^i_j|\boldsymbol{h}^i_j) = FNN(\tau^i_j, h^i_j; W^i_t).
\end{equation}
However, we need to fulfill two properties of cumulative hazard function. Firstly, it has to be a monotonically increasing function of $\tau^i_j$ and secondly, it has to be positive valued. We achieve these by maintaining positive weights and positive activation functions in the neural network~\cite{chilinski2020neural,omi2019fully}. The hazard function itself can be then obtained by differentiating the cumulative hazard function with respect to $\tau$ as 
\setlength{\abovedisplayskip}{0pt}
\setlength{\belowdisplayskip}{0pt}
\setlength{\abovedisplayshortskip}{0pt}
\setlength{\belowdisplayshortskip}{0pt}
\begin{equation}
\lambda(\tau^i_j|\boldsymbol{h}^i_j) = \frac{\partial}{\partial \tau^i_j}\Phi(\tau^i_j|\boldsymbol{h}^i_j)
\end{equation}
The log-likelihood of observing event times is defined as follows using the cumulative hazard function:
\setlength{\abovedisplayskip}{0pt}
\setlength{\belowdisplayskip}{0pt}
\setlength{\abovedisplayshortskip}{0pt}
\setlength{\belowdisplayshortskip}{0pt}
\begin{equation}
\label{eq:log_lik_nhp}
\begin{split}
& \log p(\{t^i_j\}_{j=1}^{n^i} ; W^i) = \sum_{j=1}^{n^i} \log p(t^i_j|\mathcal{H}^i_j;W^i) = \\
& \sum_{j=1}^{n^i} \big( \log (\frac{ \partial}{\partial \tau^i_j} \Phi(\tau^i_j |\boldsymbol{h}^i_{j-1}; W^i) )
 - \Phi(\tau^i_j|\boldsymbol{h}^i_{j-1}; W^i) \big) 
 \end{split}
\end{equation}
where $\tau^i_j = t^i_{j}-t^i_{j-1}$ and $W^i = \{W^i_r, W^i_t\}$ represents the combined weights associated with  RNN and  FNN.  In NHP, the weights of the networks are learnt by maximizing the likelihood given by (\ref{eq:log_lik_nhp}). The gradient of the log-likelihood function is calculated using backpropagation.

\subsection{HyperHawkes: Hypernetwork based Neural Hawkes Process}
\label{sec:HyperHawkes}
Hypernetwork is a meta-network which produces parameters used by other networks \cite{ha2017hypernetworks}. As discussed in the above section, the neural Hawkes process comprises of two building blocks - RNN and FNN. So, we use hypernetwork to produce weights for these two components. We use a feed-forward neural  network (FNN) to produce parameters $W^i = \{W_r^i, W_t^i\}$ associated with the NHP. Since the nature of the RNN parameters $W_r^i$ and FNN parameters $W_t^i$ are different, we use two different types of hypernetworks, $f_r(\cdot)$ producing $W_r^i$ and $f_t(\cdot)$ producing $W_t^i$. 
Given a sequence description $\mathbf{d}^i$, parameters for the RNN are generated as follows:
\begin{equation}
\label{eq:hrnn}
    W_r^i = f_r(\mathbf{d}^i; \theta_{fr})
\end{equation}
where $\theta_{fr}$ denotes the parameters of the hypernetwork (weight vectors of a neural network). Note that the hypernetwork parameters are the same across the sequences. The descriptor $\mathbf{d}^i$ is used to generate the sequence specific parameters. As discussed above, the cumulative hazard function is a monotonically increasing function of $\tau^i_j$ and is positive-valued. The hypernetwork which will generate parameters for cumulative hazard function has to fulfill these properties. This can be achieved when hypernetwork generates only positive weights for which we use a positive activation function. Hypernetwork $f_t(\cdot)$ for FNN can be written as:
\begin{equation}
\label{eq:hfnn}
    W_t^i = f_t(\mathbf{d}^i; \theta_{ft})
\end{equation}
where $\theta_{ft}$ denotes parameters of hypernetwork $f_t(\cdot)$.
We propose the following variants of HyperHawkes - 
\begin{itemize}
\item \textbf{HyperHawkes-FNN:} Hypernetwork is considered only for the FNN  modeling the cumulative hazard function.  
\item \textbf{HyperHawkes-FNN-RNN:} This variant uses two separate hypernetworks, one to model  the RNN modeling the history and the second to model the FNN.
\end{itemize}
Overview of the proposed architecture is shown in Fig~\ref{fig:overview}.
% Considering this, we propose two variants of the proposed approach - 
% \begin{itemize}
%     \item \textit{HyperHawkes-CHFN:} Hypernetwork is considered for the feedforward network which is modeling cumulative hazard function. Since cumulative hazard function has to take positive values, output of this hypernetwork has to be positive. 
%     \item \textit{HyperHawkes-RNN+CHFN:} This variant consists of hypernetwork for both recurrent neural network modeling history and feedforward network modeling cumulative hazard function. 
% \end{itemize}

\subsection{HyperHawkes for Zero-shot modeling}
In this section  we discuss how we employ \textit{HyperHawkes} for zero-shot event modeling. Our goal is to train the model on seen sequences $\mathcal{D}^S$ with event sequences $\mathcal{T}^s$ and task descriptors $\mathbf{d}^s$ and predict event times of unseen sequences in $\mathcal{D}^U$ given a task descriptor $\mathbf{d}^u$. We employ \textit{HyperHawkes} for performing zero-shot learning on event sequences. The central idea of the proposed approach is to predict the parameters for the neural Hawkes process for the unseen task $\mathbf{d}^u$. This is achieved using hypernetwork which considers the sequence descriptor as input and parameters for neural Hawkes process as output as discussed in the previous section. Consequently, we can get parameters for RNN ($W_r^u$) using Equation \ref{eq:hrnn} and FFN ($W_t^u$)  using Equation \ref{eq:hfnn} and use them to model the cumulative hazard function for an unseen sequence. These parameters can then be used for predicting  events in the sequence $\mathcal{T}^u$.

\textbf{Training}
We adopt a training procedure where we train hypernetwork using the maximum likelihood estimation for the NHP model. For each seen sequence $\mathcal{T}^s$ from $D^S$, we sample a mini-batch consisting of events. The sequence descriptor $\mathbf{d}^s$ of this sequence is used to generate parameters of the neural Hawkes process using the hyper-network and its parameters.  These values are then used in Equation \ref{eq:log_lik_nhp} to find the log-likelihood of the event times of the seen sequences. So, the log-likelihood described in Equation \ref{eq:log_lik_nhp} will now be:
\setlength{\abovedisplayskip}{0pt}
\setlength{\belowdisplayskip}{0pt}
\setlength{\abovedisplayshortskip}{0pt}
\setlength{\belowdisplayshortskip}{0pt}
\begin{equation}
\label{eq:log_lik_nhp_zsl}
\begin{split}
\sum_{i=1}^{N}\sum_{j=1}^{n^i}\log p(t^i_j|\mathcal{H}^i_j;(f_r(\mathbf{d}^i;\theta_{fr}),
f_t(\mathbf{d}^i;\theta_{ft}))) 
 \end{split}
\end{equation}
% where $\tau^i_j = t^i_{j}-t^i_{j-1}$ and $W^i = \{W^i_r, W^i_t\}$ represents the combined weights associated with  RNN and  FNN.  
The difference in training lies in the fact that by maximizing this log-likelihood, we will get weights of the hyperparameter network rather than the neural Hawkes process. So, we calculate $\theta_{fr}$ and $\theta_{ft}$ using the gradient of the log-likelihood function using backpropagation.

\textbf{Prediction}
For prediction of events from unseen sequence $\mathcal{T}^u$ from $D^U$, we employ our trained hypernetwork to produce weights $\{W_r^u, W_t^u\}$ for the neural Hawkes process using sequence descriptor $\mathbf{d}^u$. Neural Hawkes process uses the bisection method~\cite{omi2019fully} to predict the time of the next event. Bisection method provides  the median  $t_*$ of the predictive distribution over next event time using  the relation $\Phi(t_* - t_j^i|\boldsymbol{h}_j^i; W_r^u, W_t^u)=\log(2)$.

\subsection{HyperHawkes for Continual Learning}
In a more realistic setup of event time modeling, sequences appear one after the other and it is unrealistic to store the data and models associated with all the previous sequences. In spite of this, we need to predict correctly on these past sequences  though we have only data from new sequences. We want the NHP models to retain information from the past sequences while learning from new sequences. The standard training of the NHP model adapts them to the new sequence data (by updating parameters to optimize the loss to new sequence data) and results in forgetting what it has learnt from past sequences.  
The inability of the neural network models to retain knowledge from past data is known as catastrophic forgetting and continual learning techniques have been proposed to address this. Though it is studied in the vision community, to the best of our knowledge we could not find any work on the time-to-event prediction problem and with NHP models. 

We address the novel problem of learning the time-to-event sequences continually while retaining knowledge from past time-to-event sequences. Inspired by  \cite{von2019continual}, we use descriptor conditioned hypernetworks for continually learning from event sequences. Ideally, we want our model to remember the parameters of the neural Hawkes process for each sequence. A naive approach to achieve this is through storing and replaying over previous data, which is obviously memory expensive and unrealistic. However, \textit{HyperHawkes}, being conditioned on the sequence descriptor, can be modified to handle this problem.  The direct use of the \textit{HyperHawkes} training through \eqref{eq:log_lik_nhp_zsl} would result in hypernetworks forgetting the generation of the NHP parameters corresponding to past event sequences.  We overcome this by  incorporating  a regularization on the hypernetwork parameters such that it penalizes any change to the NHP parameters produced from old sequences.     

Given a sequence description $\mathbf{d}^s$ for the descriptor $\mathcal{T}^s$, our descriptor conditioned hypernetwork $f_r(\cdot)$ can generate parameters $W_r^s$ and $f_t(\cdot)$ can generate parameters $W_t^s$. To perform continual learning, we use regularization to penalize changes in $\{W_r^c, W_t^c\}$ generated for past sequences in order to retain information from those sequences and to learn continually. The  regularization is applied to the hypernetwork parameters while learning a new event sequence, and this prevents adaptation of the hypernetworks parameters completely to the new event sequence.  For a new event sequence $\mathcal{T}^s$ and its corresponding descriptor $\mathbf{d}^s$, the hypernetwork parameters are learnt by minimizing the following continual learning loss over events in the  sequence:
\setlength{\abovedisplayskip}{0pt}
\setlength{\belowdisplayskip}{0pt}
\setlength{\abovedisplayshortskip}{0pt}
\setlength{\belowdisplayshortskip}{0pt}
\begin{equation}
\label{eq:log_lik_nhp_cl}
\begin{split}
&\sum_{j=1}^{n^s}  -\log p(t^s_j|\mathcal{H}^s_j;(f_r(\mathbf{d}^s;\theta_{fr}), 
f_t(\mathbf{d}^s;\theta_{ft}))) \\
 & +\frac{\beta}{s-1}\sum_{c=1}^{s-1}\biggl(\parallel f_r(\mathbf{d}^c;\theta_{fr})-f_r(\mathbf{d}^c; \bar{\theta}_{fr})\!\!\parallel^2 \\
& + \parallel f_t(\mathbf{d}^c; \theta_{ft}) - f_t(\mathbf{d}^c; \bar{\theta}_{ft}) \parallel^2 \biggr)
 \end{split}
\end{equation}

where $\{\bar{\theta}_{fr}, \bar{\theta}_{ft}\}$ represents the stored hypernetwork parameters after learning until  sequence $s-1$ and $\{\theta_{fr}, \theta_{ft}\}$ represent the hypernetwork parameters learnt considering the event sequence $s$ and regularization to avoid forgetting. The regularization term ensures that the newly learnt hyper-network parameters will be able to produce the required main network parameters from the past event sequences given the sequence descriptor without forgetting and the regularization constant $\beta$ captures the importance associated with it.   So, in this way, we try to retain the information from previous sequences at a meta-level. By including a simple regularization term within the framework of \textit{HyperHawkes}, our model is capable of learning sequences continually without forgetting knowledge learnt from previous sequences. We are able to achieve this because of the use of sequence-conditioned hypernetwork on the top of neural Hawkes process, emphasizing its usefulness for continual learning over event sequences in addition to zero shot learning. 

\begin{table*}
\centering
\caption{Results for Zero-shot setup for proposed method HyperHawkes under different setups against the proposed baselines. The table considers both the variants HyperHawkes-FNN and HyperHawkes-FNN-RNN as proposed methodology for ZSL}
\label{Tab:zsl}
\tabcolsep=0.05cm
\begin{tabular}{c|c|cccc|cccc} 
\hline\hline
\multirow{2}{*}{\begin{tabular}[c]{@{}c@{}}Experimental\\Setup\end{tabular}}      & \multirow{2}{*}{Dataset} & \multicolumn{4}{c|}{MNLL}                                                                                                                                                                       & \multicolumn{4}{c}{MAE}                                                                                                                                                                         \\ 
\cline{3-10}
                                                                                  &                          & FNHP    & \begin{tabular}[c]{@{}c@{}}FNHP-\\Descriptor\end{tabular} & \begin{tabular}[c]{@{}c@{}}HyperHawkes-\\FNN\end{tabular} & \begin{tabular}[c]{@{}c@{}}HyperHawkes-\\FNN-RNN\end{tabular} & FNHP   & \begin{tabular}[c]{@{}c@{}}FNHP-\\Descriptor\end{tabular} & \begin{tabular}[c]{@{}c@{}}HyperHawkes-\\FNN\end{tabular} & \begin{tabular}[c]{@{}c@{}}HyperHawkes-\\FNN-RNN\end{tabular}  \\ 
\hline
\multirow{2}{*}{Zero-shot}                                                        & Yelp                     & -4.2023 & -3.3334                                                   & \textbf{\textbf{-5.7045}}                                 & -5.4934                                                       & 0.0025 & 0.0015                                                    & \textbf{0.0013}                                           & 0.0014                                                         \\
                                                                                  & Meme                     & -3.9011 & -2.9421                                                   & \textbf{\textbf{-5.4339}}                                 & -5.3403                                                       & 0.0056 & 0.0072                                                    & 0.0036                                                    & \textbf{0.0018}                                                \\ 
\hline\hline
\multirow{2}{*}{\begin{tabular}[c]{@{}c@{}}Generalized\\zero-shot\end{tabular}}   & Yelp                     & -4.9279 & -4.2075                                                   & -5.2473                                                   & \textbf{\textbf{-5.6869}}                                     & 0.0025 & 0.0016                                                    & 0.0021                                                    & \textbf{0.0014}                                                \\
                                                                                  & Meme                     & -4.2735 & -4.0250                                                   & -5.2475                                                   & \textbf{\textbf{-5.2922}}                                     & 0.0036 & 0.0033                                                    & 0.0030                                                    & \textbf{0.0026}                                                \\ 
\hline\hline
\multirow{2}{*}{\begin{tabular}[c]{@{}c@{}}Standard\\event modeling\end{tabular}} & Yelp                     & -4.2700 & -3.8543                                                   & -4.9030                                                   & \textbf{\textbf{-4.9475}}                                     & 0.0047 & 0.0042                                                    & \textbf{0.0024}                                           & 0.0025                                                         \\
                                                                                  & Meme                     & -4.6991 & -2.9633                                                   & -2.5325                                                   & \textbf{-4.8796}                                                      & 0.0046 & 0.0072                                                    & 0.0078                                                    & \textbf{0.0027}                                                \\
\hline\hline
\end{tabular}
\end{table*}

\section{Experiments}

\subsection{Datasets}
Due to paucity of standard datasets for event modeling tasks which contain meta descriptions as well, we use the following two datasets: \textbf{1)Yelp:}\footnote[1]{https://www.kaggle.com/datasets/yelp-dataset/}: This is a dataset comprising of business information and their check-in information. Each business is associated with 82 attributes like \textit{Wheelchair Accessible, Accepts Insurance, By Appointment Only, Business Category, Business Timings} etc. Also, they are associated with latitude-longitude pairs. Moreover, these businesses are associated with a fine-grained category. For higher granularity, we convert them into 22 broad categories using the hierarchy mentioned in their website \footnote[2]{https://www.yelp.com/developers/documentation/v3/all\_category\_list}. Using these attributes, we create a vector of length 1229 representing a business. This vector acts as a descriptor of the business. We select businesses with more than 5000 check-ins. For continual learning, we have considered another sample of dataset consisting of business with more than 10k checkins, hence considering 26 sequences of business. This is done to reduce the number of sequences for better visualization of performance of each sequence. \textbf{2) Meme:}\footnote[3]{https://snap.stanford.edu/data/memetracker9.html}: This dataset\cite{leskovec2009meme} tracks the popular phrases and quotes which appear appear most frequently over time in news media and blogs. Each meme is associated with the content and timestamps when they were quoted in the media. We select top 200 english phrases and doc2vec representation of the meme content is considered as the descriptor of length 100. We have considered memes from April, 2009 with an average number of events as 970. 

\subsection{Baselines}
To the best of our knowledge, the proposed problem statement is the first work along this direction. Therefore, we propose our own baselines as - \textbf{1) FNHP:} This includes fully neural Hawkes process \cite{omi2019fully}. This approach doesn't incorporate the sequence descriptor. \textbf{2) FNHP-Descriptor:} In this variant, we use concatenated descriptor and time as input to RNN and FNN.  For Continual learning setup, we compare against HyperHawkes without any regularization as baseline. 
% \begin{itemize}
%     \item \textbf{FNHP:} This includes fully neural Hawkes process \cite{omi2019fully}.
%     \item \textbf{MetaFNHP:} In this variant, we use concatenated meta descriptor and time as input to RNN and FNN. 
%  \end{itemize}  

\subsection{Implementation Details}
For zero-shot setup, we perform a 60-20-20 split where 60\% of sequences are considered as seen sequence and 20\% for validation unseen sequence and rest 20\% as test unseen sequence. We have used a single layer with 32 units for hypernetwork where we use softplus activation function for modeling cumulative hazard function. For the neural Hawkes process, we consider a  recurrent neural network with one layer and 16 units and  2-layer feed-forward neural network with 16 units in each layer. We use Adam optimizer with learning rate, $\beta_1$ and $\beta_2$ as 0.0001, 0.90 and 0.99 for the reported results. We perform single step lookahead prediction where we use actual time of occurrence of events as past events for the historical information to predict future events. We have performed all the experiments on Intel(R) Xeon(R) Silver 4208 CPU @ 2.10GHz, GeForce RTX 2080 Ti GPU and 128 GB RAM. We have implemented our code in Tensorflow 2.2.0 \cite{tensorflow2015-whitepaper}. For continual learning setup, we test on regularization parameters in range [0.001-0.9]. For Yelp, reported results are for $\beta$ is set as 0.5 for HyperHawkes-FNN, 0.6 for HyperHawkes-FNN-RNN and for Meme $\beta$ is 0.6 for HyperHawkes-FNN and 0.1 for  HyperHawkes-FNN-RNN. More implementation details and hyperparameter settings are presented in the supplementary.

\begin{table*}
\centering
\caption{Results for Continual Learning by comparing the performance of the proposed HyperHawkes with regularization against under HyperHawkes without regularization (Lower MNLL and MAE indicates better performance) 
% (Without CL corresponds to the case when $\beta$ from Equation \ref{eq:log_lik_nhp_cl} is set to 0)
}
\label{Tab:cl}
\tabcolsep=0.2cm
\begin{tabular}{c|cc|cc||cc|cc} 
\hhline{=====:t:====}
\multirow{3}{*}{Dataset} & \multicolumn{4}{c||}{HyperHawkes-FNN}                                                          & \multicolumn{4}{c}{HyperHawkes-FNN-RNN}                                                                           \\ 
\cline{2-9}
                         & \multicolumn{2}{c|}{MNLL}                               & \multicolumn{2}{c||}{MAE}            & \multicolumn{2}{c|}{MNLL}                               & \multicolumn{2}{c}{MAE}                                 \\ 
\cline{2-9}
                         & WithoutCL & WithCL                                      & WithoutCL & WithCL                   & WithoutCL & WithCL                                      & WithoutCL & WithCL                                      \\ 
\hline
Yelp                     & -4.8627   & \textbf{\textbf{\textbf{\textbf{-5.6928}}}} & 0.00159    & \textbf{\textbf{0.00149}} & -5.2629   & \textbf{\textbf{\textbf{\textbf{-5.7727}}}} & 0.00152    & \textbf{\textbf{\textbf{\textbf{0.00150}}}}  \\
Meme                     & -2.8550   & \textbf{\textbf{-5.1462}}                   & 0.00548    & \textbf{0.00471}          & -3.8254   & \textbf{\textbf{-5.1192}}                   & 0.00508    & \textbf{\textbf{0.00471}}                    \\
\hhline{=====:b:====}
\end{tabular}
\end{table*}

      \begin{figure*}[t]
  \centering
  \caption{Plots displaying the performance of HyperHawkes for Continual Learning for the variant HyperHawkes-FNN and HyperHawkes-FNN-RNN \ref{fig:avg_mnll_yelp}) Average MNLL over previous sequences for Yelp \ref{fig:cl_chfn_rnn_averaged_mae_meme}) Average MAE over previous sequences for Meme \ref{fig:mnll_yelp}) MNLL for each sequence for Yelp HyperHawkes-FNN  \ref{fig:ablation}) MNLL for different values of $\beta$ in Eq. \ref{eq:log_lik_nhp_cl} for HyperHawkes-FNN in Yelp and Meme }
 \label{fig:cl_plots}
  \subfloat[] {\includegraphics[scale=0.21]{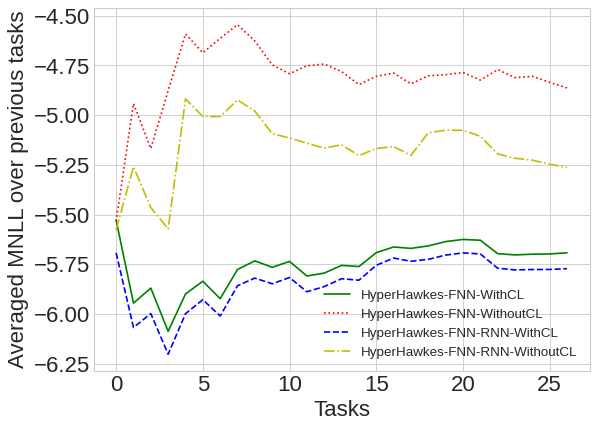} \label{fig:avg_mnll_yelp}} \quad
  \subfloat[]{\includegraphics[scale=0.21]{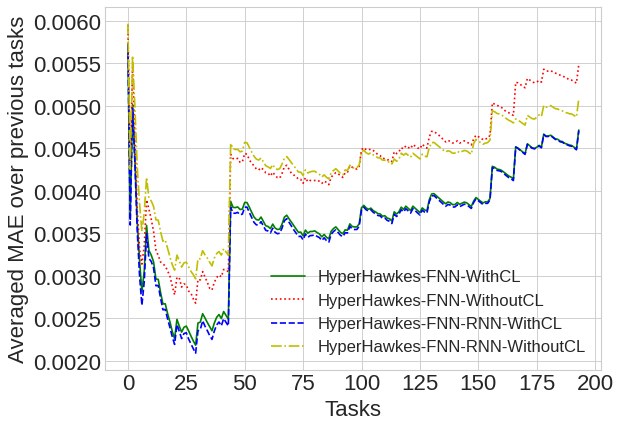} \label{fig:cl_chfn_rnn_averaged_mae_meme}} 
    \quad
  \subfloat[]{\includegraphics[scale=0.21]{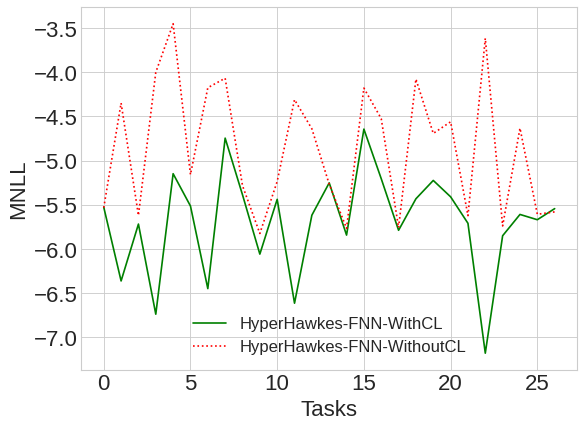}  \label{fig:mnll_yelp}}
  \quad
  \subfloat[]{\includegraphics[scale=0.21]{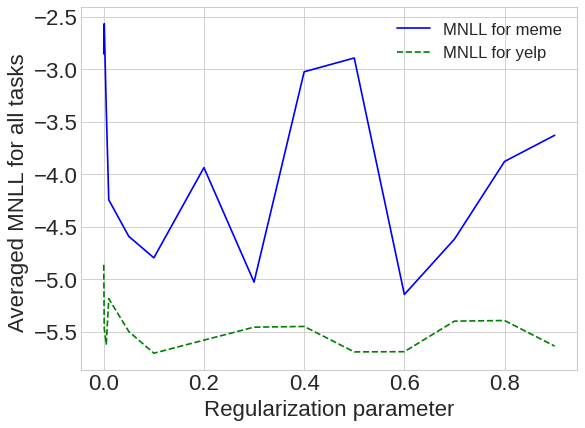} \label{fig:ablation}} 
  \end{figure*}

\subsection{Experimental Setup}
We consider these experimental setups to evaluate the performance of our model - 
\textbf{1) Zero-Shot:} Training is done on seen sequences and testing is done on unseen sequences.  \textbf{2) Generalized Zero-Shot:} In this, testing is done by randomly sampling 20\% of events from seen sequences and unseen sequences.  \textbf{3) Standard Event Modeling:} Training is done on the first 70\% of the events from all sequences. Testing is done on the last 20\% events for unseen sequences. 
Mean negative log-likelihood (MNLL) and mean absolute error (MAE) are considered as evaluation metrics for both zero-shot and continual learning setup. Lower MNLL and MAE indicates better performance.
% \begin{itemize}
%     \item \textbf{Zero-Shot:} Training is done on seen sequences and testing is done on unseen sequences.  
%     \item \textbf{Generalized Zero-Shot:} In this, testing is done by randomly sampling 20\% of events from seen sequences and unseen sequences. 
%     \item \textbf{Standard Event Modeling:} Training is done on the first 70\% of the events from seen sequences. Testing is done on the last 20\% events for unseen sequences. 
%  \end{itemize}   
    
    % In this case, we train based on the events on the first 70\% of the events from all the businesses and while prediction, we predict the last 20\% of the events from all the events.
    % \item \textbf{Setup 4: Data Efficiency Zero-Shot} We check the model performance when the number of tasks is less.

\section{Results and Analysis}
\subsection{Zero-Shot Learning}
Results for zero-shot setup are presented in Table \ref{Tab:zsl}. A better model is expected to have lower MNLL and MAE. We can observe that the proposed method HyperHawkes performs better than the baselines in terms of both evaluation metrics MNLL and MAE. We can observe that zero-shot setup, which consists of predictive performance for unseen sequences, exhibits significantly lower MNLL and MAE for HyperHawkes-FNN as compared to FNHP and FNHP-Descriptor. Also, HyperHawkes-FNN-RNN yields lower MAE for Meme dataset, however, MNLL for both the variants of the proposed methods is close. Comparing the results of generalized zero-shot setup where we consider instances from seen sequences as well as unseen sequences, we can observe that HyperHawkes-FNN-RNN performs better for both the datasets. The final section of the table discusses results for standard event modeling where we test on the last 20\% of the events of unseen sequences. HyperHawkes-FNN-RNN performs well here as well except for MAE for Yelp dataset. We can also observe that HyperHawkes-FNN-RNN performs slightly better for generalized zero-shot and standard event modeling setup. Moreover, it is interesting to note that inclusion of sequence descriptors within FNHP in FNHP-Descriptor has not helped much in prediction of unseen sequences. In fact, in various cases, it has performed worse than FNHP itself. This confirms the necessity of an event sequence specific framework which can predict better for unseen sequences as well. Therefore, our results indicate that the proposed variants of HyperHawkes perform consistently and significantly better than the baselines for both the datasets.

\subsection{Continual Learning}

Table \ref{Tab:cl} presents the averaged results over all tasks by enabling HyperHawkes for continual learning. We can observe that averaged performance for the proposed model is better than the case when no regularization is incorporated. Also, we can observe that both the proposed variants perform better than the model without using regularization (corresponds to the case when $\beta$ from Equation \ref{eq:log_lik_nhp_cl} is set to 0). Hence the use of regularization within the framework of HyperHawkes supports that proposed method can avoid catastrophic forgetting. Fig \ref{fig:cl_plots} displays sequence-wise performance for both the datasets for the proposed variants HyperHawkes-FNN and HyperHawkes-FNN-RNN.\ref{fig:avg_mnll_yelp}) displays average MNLL over previous sequences for both the models for Yelp. This shows that while training the model without regularization over new sequences, the network is unable to retain information learnt from previous sequences, hence MNLL increases as we train new sequences. However, with the use of regularization, we can avoid catastrophic forgetting, hence having lower MNLL for successive tasks. Similar behavior is observed by \ref{fig:cl_chfn_rnn_averaged_mae_meme}) as well which displays average MAE over previous sequences for both the variants, with and without CL for Meme. So, this corroborates that use of regularization with HyperHawkes can help in backward transfer. \ref{fig:mnll_yelp}) shows MNLL for each sequence using the model HyperHawkes-FNN-RNN with regularization. MNLL for the model with regularization is having lower MNLL as compared to the model without any regularization. This essentially reflects that the proposed model is able to forward transfer the knowledge learnt from previous sequences as well. So, the model is able to perform forward and backward transfer, which are important continual learning desiderata. \ref{fig:ablation}) displays the effect of various regularization parameters for Yelp and Meme dataset for HyperHawkes-FNN. A possible explanation could be for Meme for $\beta$, the model is not able to learn from previous sequences and for large $\beta$ might not be able to learn from new sequences. To conclude, presented results suggest that proposed framework can aid in avoiding catastrophic forgetting while learning continually.

\section{Conclusion}
In this work, we address two novel and practical limitations for time-to-event modeling. Firstly, we address zero-shot event modeling for predicting time of unseen events. Secondly, we propose an approach for continual learning for time-to-event modeling where we learn when event sequences continually and the model learns while retaining previous knowledge. To address both of these issues, we propose HyperHawkes, a descriptor conditioned hypernetwork based neural Hawkes process which can generate event sequence specific parameters. The proposed approach can predict the time of occurrences of events from unseen sequences, hence performing zero-shot time-to-event modeling. Subsequently, we augment \textit{HyperHawkes} with regularization which can aid in learning time-to-event sequences continually by avoiding catastrophic forgetting. Our experiments on two real-world datasets demonstrate the effectiveness of the proposed approach for both the issues. In this way, we augment the ability of the neural Hawkes process to perform two unexplored and practical tasks of zero-shot and continually learning time-to-event modeling.

\bibliography{aaai23}

\end{document}